\documentclass[journal,twoside,web]{ieeecolor}

\usepackage{cite}
\usepackage{amsmath,amssymb,amsfonts}
\usepackage{algorithmic}
\usepackage{graphicx}
\setkeys{Gin}{draft=false}
\usepackage{textcomp}
\usepackage{multirow}
\usepackage{siunitx}
\usepackage{textcomp}
\usepackage{verbatim}
\usepackage[normalem]{ulem} 
\usepackage{amssymb}
\usepackage{import}
\usepackage{graphicx} 
\usepackage{booktabs}
\usepackage{amsmath}
\usepackage[x11names]{xcolor}
\graphicspath{{fig/}}

\providecommand{\mytitle}{Bladder Vessel Segmentation using a Hybrid Attention-Convolution Framework}
\def\logoname{fig/logo}
\definecolor{subsectioncolor}{rgb}{0,0,0}
\def\BibTeX{{\rm B\kern-.05em{\sc i\kern-.025em b}\kern-.08em
    T\kern-.1667em\lower.7ex\hbox{E}\kern-.125emX}}
\newcommand{\cv}[1]{\textcolor{blue}{#1}}

\newcommand{\meanstd}[2]{#1\,±\,#2}
\newcommand{\meanstdpercent}[2]{$#1\,\pm\,#2\,\%$}

\begin{document}

\definecolor{WineRed}{RGB}{200,0,0}
\newcommand{\tbd}[1]{\textcolor{WineRed}{[tbd] #1}}
\definecolor{Orange}{RGB}{255, 150, 79}
\newcommand{\question}[1]{\textcolor{Orange}{[question] #1 ??}}
\newcommand{\up}{\ensuremath{\uparrow}}
\newcommand{\down}{\ensuremath{\downarrow}}

\newcommand{\accunet}{0.93}
\newcommand{\precunet}{0.47}
\newcommand{\recunet}{0.75}
\newcommand{\foneunet}{0.58}
\newcommand{\iouunet}{0.41}
\newcommand{\diceunet}{0.58}
\newcommand{\cldiceunet}{0.30}
\newcommand{\avgfragunet}{262.45}
\newcommand{\numcompunet}{106.50}
\newcommand{\bettiunet}{67.38}

\newcommand{\accunetpp}{0.92}
\newcommand{\precunetpp}{0.45}
\newcommand{\recunetpp}{0.80}
\newcommand{\foneunetpp}{0.57}
\newcommand{\iouunetpp}{0.40}
\newcommand{\diceunetpp}{0.57}
\newcommand{\cldiceunetpp}{0.22}
\newcommand{\avgfragunetpp}{567.64}
\newcommand{\numcompunetpp}{55.50}
\newcommand{\bettiunetpp}{36.12}

\newcommand{\accsaunet}{0.94}
\newcommand{\precsaunet}{0.58}
\newcommand{\recsaunet}{0.67}
\newcommand{\fonesaunet}{0.62}
\newcommand{\iousaunet}{0.45}
\newcommand{\dicesaunet}{0.62}
\newcommand{\cldicesaunet}{0.49}
\newcommand{\avgfragsaunet}{676.85}
\newcommand{\numcompsaunet}{31.50}
\newcommand{\bettisaunet}{25.62}

\newcommand{\accswinunet}{0.94}
\newcommand{\precswinunet}{0.57}
\newcommand{\recswinunet}{0.38}
\newcommand{\foneswinunet}{0.45}
\newcommand{\iouswinunet}{0.30}
\newcommand{\diceswinunet}{0.45}
\newcommand{\cldiceswinunet}{0.33}
\newcommand{\avgfragswinunet}{94.49}
\newcommand{\numcompswinunet}{127.20}
\newcommand{\bettiswinunet}{90.60}

\newcommand{\acchac}{0.94}
\newcommand{\prechac}{0.61}
\newcommand{\rechac}{0.59}
\newcommand{\fonehac}{tbd}
\newcommand{\iouhac}{0.43}
\newcommand{\dicehac}{0.60}
\newcommand{\cldicehac}{0.66}

\providecommand{\blavesOverallRatioMean}{69.41}\providecommand{\blavesOverallRatioStd}{6.16}
\providecommand{\blavesOverallRIMean}{136.45}\providecommand{\blavesOverallRIStd}{8.90}
\providecommand{\blavesOverallRCCMean}{0.02}\providecommand{\blavesOverallRCCStd}{0.07}
\providecommand{\blavesOverallCSIMean}{16.12}\providecommand{\blavesOverallCSIStd}{11.22}
\providecommand{\blavesOverallFGIMean}{3306.25}\providecommand{\blavesOverallFGIStd}{934.69}
\providecommand{\blavesOverallCVMean}{0.25}\providecommand{\blavesOverallCVStd}{0.09}
\providecommand{\blavesOverallVIMean}{0.37}\providecommand{\blavesOverallVIStd}{0.09}
\providecommand{\blavesVesselRatioMean}{7.51}\providecommand{\blavesVesselRatioStd}{2.37}
\providecommand{\blavesVesselRIMean}{138.79}\providecommand{\blavesVesselRIStd}{7.80}
\providecommand{\blavesVesselTIMean}{1.19}\providecommand{\blavesVesselTIStd}{0.09}
\providecommand{\blavesVesselBDMean}{8.75}\providecommand{\blavesVesselBDStd}{2.78}
\providecommand{\blavesVesselCCMean}{1.07}\providecommand{\blavesVesselCCStd}{0.04}
\providecommand{\blavesVesselFGIMean}{4904.26}\providecommand{\blavesVesselFGIStd}{1970.95}
\providecommand{\blavesVesselCVMean}{0.22}\providecommand{\blavesVesselCVStd}{0.08}
\providecommand{\blavesVesselVIMean}{0.32}\providecommand{\blavesVesselVIStd}{0.09}
\providecommand{\blavesBGRatioMean}{62.13}\providecommand{\blavesBGRatioStd}{6.20}
\providecommand{\blavesBGRIMean}{136.06}\providecommand{\blavesBGRIStd}{9.31}
\providecommand{\blavesBGFGIMean}{3040.07}\providecommand{\blavesBGFGIStd}{684.82}
\providecommand{\blavesBGCVMean}{0.26}\providecommand{\blavesBGCVStd}{0.09}
\providecommand{\blavesBGVIMean}{0.38}\providecommand{\blavesBGVIStd}{0.09}
\providecommand{\driveOverallRatioMean}{65.07}\providecommand{\driveOverallRatioStd}{0.43}
\providecommand{\driveOverallRIMean}{190.06}\providecommand{\driveOverallRIStd}{32.82}
\providecommand{\driveOverallRCCMean}{-0.02}\providecommand{\driveOverallRCCStd}{0.03}
\providecommand{\driveOverallCSIMean}{12.41}\providecommand{\driveOverallCSIStd}{3.89}
\providecommand{\driveOverallFGIMean}{407.93}\providecommand{\driveOverallFGIStd}{85.60}
\providecommand{\driveOverallCVMean}{0.10}\providecommand{\driveOverallCVStd}{0.05}
\providecommand{\driveOverallVIMean}{0.13}\providecommand{\driveOverallVIStd}{0.11}
\providecommand{\driveVesselRatioMean}{8.69}\providecommand{\driveVesselRatioStd}{1.39}
\providecommand{\driveVesselRIMean}{187.25}\providecommand{\driveVesselRIStd}{33.58}
\providecommand{\driveVesselTIMean}{1.10}\providecommand{\driveVesselTIStd}{0.01}
\providecommand{\driveVesselBDMean}{7.53}\providecommand{\driveVesselBDStd}{1.28}
\providecommand{\driveVesselCCMean}{1.06}\providecommand{\driveVesselCCStd}{0.02}
\providecommand{\driveVesselFGIMean}{866.01}\providecommand{\driveVesselFGIStd}{219.16}
\providecommand{\driveVesselCVMean}{0.09}\providecommand{\driveVesselCVStd}{0.04}
\providecommand{\driveVesselVIMean}{0.16}\providecommand{\driveVesselVIStd}{0.10}
\providecommand{\driveBGRatioMean}{56.59}\providecommand{\driveBGRatioStd}{1.45}
\providecommand{\driveBGRIMean}{190.31}\providecommand{\driveBGRIStd}{32.77}
\providecommand{\driveBGFGIMean}{329.86}\providecommand{\driveBGFGIStd}{63.16}
\providecommand{\driveBGCVMean}{0.10}\providecommand{\driveBGCVStd}{0.05}
\providecommand{\driveBGVIMean}{0.12}\providecommand{\driveBGVIStd}{0.11}

\title{\mytitle}
\author{Franziska Krauß$^{1}$,
	    Matthias Ege$^{1}$,
        Zoltan Lovasz$^{1}$, 
        Albrecht Bartz-Schmidt$^{2}$, 
        Igor Tsaur$^{2}$,
        Oliver Sawodny$^{1}$, and
        Carina Veil$^{3}$
\thanks{*This work was conducted in the framework of the Graduate School
2543/2 ”Intraoperative Multisensory Tissue Differentiation in Oncology”
(project ID 40947457) funded by the German Research Foundation (DFG-
Deutsche Forschungsgemeinschaft).}
\thanks{Franziska Krauß, Matthias Ege, Zoltan Lovasz, and Oliver Sawodny are with the Institute for System Dynamics in the University of Stuttgart, Stuttgart, Germany (e-mail: franziska.krauss@isys.uni-stuttgart.de).}
\thanks{Igor Tsauer and Albrecht Bartz-Schmidt are with the Department of Urology, University Hospital Tübingen, 72076 Tübingen, Germany. 
}
\thanks{Carina Veil is with the Department of Mechanical Engineering, Stanford University, Stanford, CA 94305, USA.
}
}

\maketitle

\begin{abstract}
Urinary bladder cancer surveillance requires tracking tumor sites across repeated interventions, yet the deformable and hollow bladder lacks stable landmarks for orientation. While blood vessels visible during endoscopy offer a patient-specific ``vascular fingerprint" for navigation, automated segmentation is challenged by imperfect endoscopic data, including sparse labels, artifacts like bubbles or variable lighting, continuous deformation, and mucosal folds that mimic vessels. State-of-the-art vessel segmentation methods often fail to address these domain-specific complexities. We introduce a Hybrid Attention-Convolution (HAC) architecture that combines Transformers to capture global vessel topology prior with a CNN that learns a residual refinement map to precisely recover thin-vessel details. To prioritize structural connectivity, the Transformer is trained on optimized ground truth data that exclude short and terminal branches. Furthermore, to address data scarcity, we employ a physics-aware pretraining, that is a self-supervised strategy using clinically grounded augmentations on unlabeled data.
Evaluated on the BlaVeS dataset, consisting of endoscopic video frames, our approach achieves high accuracy (0.94) and superior precision (0.61) and clDice (0.66) compared to state-of-the-art medical segmentation models. Crucially, our method successfully suppresses false positives from mucosal folds that dynamically appear and vanish as the bladder fills and empties during surgery. Hence, HAC provides the reliable structural stability required for clinical navigation.
\end{abstract}



\section{Introduction}\label{ch:introduction}
\begin{figure*}[t]
    \centering
    \includegraphics[width=1\linewidth]{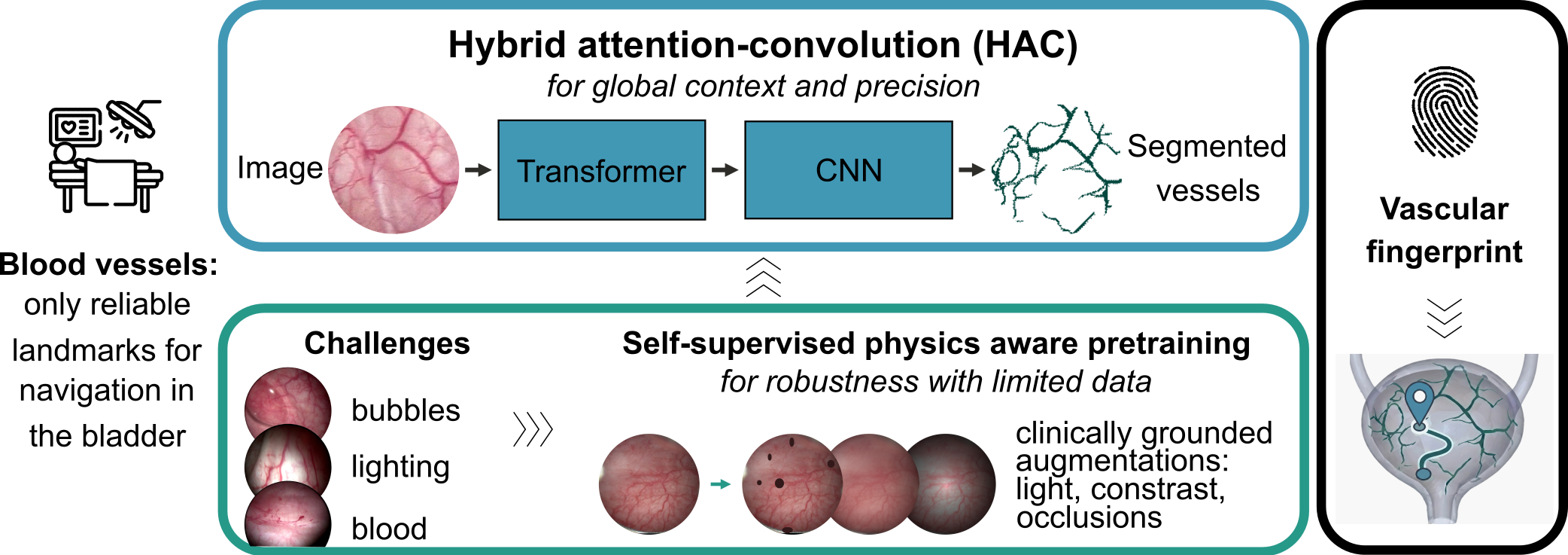}
    \caption{Framework Overview. Blood vessels serve as a patient-specific vascular fingerprint for navigation. We utilize self-supervised physics-aware pre-training to handle cystoscopic challenges (e.g., bubbles, variable lighting) and a Hybrid Attention-Convolution (HAC) architecture to combine global context with fine-scale precision for robust segmentation.} 
    \label{fig:graphical_abstract}
\end{figure*}
Bladder cancer ranks among the most common malignancies worldwide. In early stages of non-muscle-invasive bladder cancer (NMIBC), the tumors are confined to the inner layer of the bladder wall and electrosurgically removed during endoscopic interventions through the urethra (\emph{cystoscopy}). 

The treatment of NMIBC is associated with high recurrence rates of up to \qty{70}{\percent} within five years \cite{Jubber2023, Abbas2025}, and tight surveillance with frequent and repeated cystoscopies is required to avoid recurrence or progression to muscle-invasive stages. A major difficulty of follow-up procedures lies in identifying and monitoring suspicious or previously treated areas over repeated interventions, especially due to the visually homogeneous mucosa and the bladder's continuous deformation during surgery.
\textbf{Patient-specific digital twins}, that is, computational models that maintain geometric and anatomical correspondence across examinations, represent a promising approach to address this challenge; however, they require robust, deformation-invariant anatomical landmarks that are difficult to find in the soft and hollow organ. 
The blood vessels in the bladder visible during cystoscopy offer a promising solution, as they form a patient-specific branching topology that functions as a \textbf{vascular fingerprint} of the bladder. Once segmented, the vessel structure enables reliable localization throughout the organ during surgery.
As the bladder wall provides no other stable geometric references, the vascular topology effectively is the only consistent anatomical landmark for navigation,
yet its use is complicated by strong organ deformation and severe imaging variability, including nonlinear vignetting, specular reflections, and diverse tissue coloration. These factors, combined with a sparsity of annotated datasets in this underexplored domain, result in insufficient and imperfect supervision for segmentation models.

\subsection{Related Work} 
Cystoscopic image segmentation has focused either on whole-bladder delineation in abdominal scans or on tumor detection and segmentation \cite{Lubbad2024, Sudhi2023}. Apart from our previous works, where we have set the foundation for vascular graph extraction \cite{Schuele2022a} and established the first hand-labeled dataset for bladder vessel segmentation \cite{Krauss2025}, cystoscopic vessel segmentation has not been addressed.

\textbf{Blood vessel segmentation} in general is most mature in the application of the retina, with benchmark datasets DRIVE \cite{Staal2004}, STARE \cite{Hoover2000}, and CHASE\_DB \cite{Fraz2012}, which are, however, acquired under standardized and quasi-planar imaging conditions. \textit{In this mature retinal setting}, early methods employed hand-crafted features such as matched filters \cite{Chaudhuri1989}, morphological operators \cite{Zana2001}, and multi-scale Gaussian kernels \cite{Soares2006}. After the introduction of U-Net \cite{Ronneberger2015}, the approach shifted to deep learning architectures \cite{Alom2018}, and extensions include dense skip pathways between the encoder and decoder sub-network \cite{Zhou2018} or spatial attention gates to suppress irrelevant features \cite{Oktay2018}. Topological losses prevent vessel fragmentation by enforcing centerline connectivity \cite{Shit2021}. 
Beyond retinal imaging, vessel segmentation has also been explored in coronary angiography \cite{Nasr-Esfahani2016} and cerebral vasculature \cite{Ajam2017}.
Yet, all these advances rely on \emph{controlled imaging conditions} with minimal deformation and consistent contrast, unlike deformable cystoscopy.

The complex shapes and fine details of vessel segmentation are difficult to capture for traditional deep learning methods alone. \textbf{Vision transformers} can consider the entire image at once, enabling the model to understand global context and relationships that help improve segmentation accuracy.
The core mechanism for this global understanding is the attention mechanism, which allows the model to selectively focus on relevant image regions and capture long-range dependencies.
Convolutional attention modules \cite{Woo2018} enhance CNNs by recalibrating feature maps, while vision transformers (ViTs) \cite{Dosovitskiy2020} use self-attention to capture global dependencies, proving effective for medical image segmentation \cite{Chen2021, Cao2022}. Hybrid architectures that combine transformer encoders with CNN decoders \cite{Wang2022, Hatamizadeh2022} exploit both global context modeling and fine-grained localization. Specifically, in retinal blood vessel segmentation, transformer-based methods use self-attention for global context and cross-attention to highlight vessels while suppressing background noise \cite{Vaswani2017, Chen2021, Petit2021}.

While ViTs are powerful, they require large amounts of labeled data, which is typically sparse in medical imaging, including cystoscopy. This motivates \textbf{self-supervised learning} strategies which first learn robust features from unlabeled images before fine-tuning on limited labels. Methods like contrastive learning \cite{Chen2020}, masked autoencoders \cite{He2022}, and consistency regularization \cite{Laine2017}, enable domain adaptation with minimal target-domain labels. 
In medical imaging, self-supervised pretraining on large unlabeled datasets \cite{Zhou2021, Azizi2021} improves generalization.
Existing methods also use domain randomization \cite{Tobin2017} and photometric augmentation \cite{Buslaev2020} to simulate variability, but treat these transformations as random i.i.d. samples rather than modeling clinically relevant deformations like those in cystoscopy.

\subsection{Problem Statement}
The blood vessels visible during cystoscopy form a unique vascular fingerprint that enables reliable navigation and tracking of suspicious areas across repeated procedures. However, vessel segmentation in cystoscopy faces unique challenges absent in retinal benchmarks:
\begin{itemize}
    \item \textbf{Intraoperative imaging artifacts}: Bubbles, specular reflections, shadows, surgical tools, non-uniform lighting, and extreme procedural variability without standardized conditions.
    \item \textbf{Multi-layer anatomy}: Vessels span multiple bladder wall layers at varying depths with scars and invaginations. 
    This superposition causes vessels to appear in varying colors and results in frames containing regions from sharply focused to heavily blurred, often leading to apparent vessel discontinuities. 
    \item \textbf{Sparse, imperfect labels}: No \emph{large/public} annotated datasets exist beyond our prior work, and labels remain scarce, imbalanced, and noisy.
    \item \textbf{Continuous deformation}: Non-rigid bladder motion disrupts topology. Emptying the bladder during the surgery induces folds that morphologically mimic vascular structures. This leads to vessel predictions that disappear in the emptied state and thereby reduce landmark stability.
    \item \textbf{Architectural limitations}: While CNNs miss the global vascular context, ViTs sacrifice resolution for thin vessels, and generic self-supervision ignores clinically relevant cystoscopic variability.  
\end{itemize}

\subsection{Contributions}

We tackle the challenge of navigating the bladder during (repeated) cystoscopies by making vessel segmentation reliable even with imperfect cystoscopic data. As shown in Figure \ref{fig:graphical_abstract}, our approach combines two key contributions:
\begin{itemize}
    \item \textbf{Hybrid attention-convolution (HAC)} architecture: Transformers capture the global vessel network, while CNNs precisely outline thin vessel edges.
    \item \textbf{Physics-aware pre-training}:  Self-supervised learning uses clinically grounded augmentations (e.g., bubbles, vignetting) to model cystoscopic geometry and artifacts.
\end{itemize}
Together, HAC provides navigation-ready vascular maps while the pre-training ensures reliability under real surgery conditions. This paper is organized as follows: Section~\ref{sec:methods} describes the BlaVeS dataset, presents the HAC architecture with its attention backbone and U-Net refinement, and provides our three-stage training protocol (Unsupervised
physics-aware pre-training, global topology learning, and HAC refinement). Section~\ref{sec:results} evaluates HAC against state-of-the-art methods and with a qualitative analysis. Section~\ref{sec:discussion} discusses the results and Section~\ref{sec:conclusion} concludes.

\section{Methods}\label{sec:methods}

\subsection{Dataset}
BlaVeS \cite{Blaves25, Krauss2025} is currently the only publicly available vessel‑annotated cystoscopic dataset. It contains 50 hand-labeled frames extracted from four video sequences\footnote{All videos were recorded at the Department of Urology, University Hospital Tübingen, and written consent was obtained from all participants.}. (512×512 resolution) and serves as the primary training set for the proposed network. The bladder's vasculature is highly tortuous, with frequent branching and diameters ranging from 10 to \qty{250}{\micro\meter} \cite{Miodonski1999, Andersson2017}. Cystoscopic images capture a superposition of these vascular layers across the bladder's distinct wall structures, often complicated by prominent mucosal folds. This anatomical complexity results in frames containing regions ranging from sharply focused to heavily blurred. Intraoperative imaging conditions further add to these challenges through nonuniform illumination, pronounced brightness, and color variations \cite{Chang2024}. To point out the challenges of vessel segmentation in the bladder’s endoscopic environment, BlaVeS is systematically analyzed and compared to the well-established retinal vessel dataset DRIVE \cite{Staal2004} in the Appendix.

\subsection{Hybrid Attention-Convolution Network}
Bladder vessel segmentation requires balancing two objectives: (i) global connectivity for navigation orientation and (ii) fine-scale branches for precise localization. Attention mechanisms capture long-range dependencies but lose resolution on thin vessels; CNNs excel at thin structures but miss global context. \textbf{HAC} couples these strengths: the attention backbone learns a vascular topology prior, which is concatenated with the original image and forwarded to a U-Net decoder that predicts a residual refinement map which is added to this prior to sharpen boundaries and recover small-scale details.

\subsubsection{Attention Backbone for Global Vessel Network}
The attention backbone is designed to extract a robust vascular skeleton without overfitting to mucosal folds or scars. The relevant parameters are summed up in Table~\ref{tab:hac_params}.

\textbf{Fine-Scale Patch Embedding.}
Let $I \in \mathbb{R}^{H\times W\times 3}$ be an RGB cystoscopic image. To balance resolution and performance, we partition the image into non-overlapping $p\times p$ patches with $ p=8$. 

This yields a sequence of $N = \frac{H}{p} \times \frac{W}{p} {=} 4 096$ patches. Each patch is projected to an embedding dimension $d=384$ via a learned convolutional layer. The resulting spatial feature map $ \mathbf{Z}^{(0)}~\in~\mathbb{R}^{(H/p) \times (W/p) \times d}$ is flattened to produce a sequence of $N$ tokens $ \mathbf{z}^{(0)} \in \mathbb{R}^{N \times d}$.

\textbf{Transformer Encoder.} 
The token sequence $\mathbf{z}^{(0)}$ is processed through $L=8$ transformer encoder blocks. Each block applies multi-head self-attention (MSA) followed by a feed-forward network (MLP), both with pre-layer normalization (LN) and residual connections
\begin{align}
    \mathbf{z}_{\ell}' &= \text{MSA}(\text{LN}(\mathbf{z}_{\ell-1})) + \mathbf{z}_{\ell-1}, \label{eq:transformer_attn} \\
    \mathbf{z}_{\ell} &= \text{MLP}(\text{LN}(\mathbf{z}_{\ell}')) + \mathbf{z}_{\ell}', \label{eq:transformer_mlp}
\end{align}
where $\ell \in \{1, \ldots, L\}$ is the block index. The MLP consists of two linear layers with GELU activation and expansion ratio $r{=}4.0$. The MSA mechanism enables each token to aggregate information from all other tokens in the sequence. We utilize $h=4$ parallel attention heads, where each head $i$ computes scaled dot-product attention
\begin{equation}
    \text{head}_i = \text{softmax}\left(\frac{\mathbf{Q}_i \mathbf{K}_i^\top}{\sqrt{d_k}}\right) \mathbf{V}_i,
\end{equation}
with queries $\mathbf{Q}_i = \mathbf{Z}\mathbf{W}_i^Q$, keys $\mathbf{K}_i = \mathbf{Z}\mathbf{W}_i^K$, and values $\mathbf{V}_i = \mathbf{Z}\mathbf{W}_i^V$, where $\mathbf{W}_i^Q, \mathbf{W}_i^K, \mathbf{W}_i^V \in \mathbb{R}^{d \times d_k}$ are learned projection matrices. The per-head dimension is $d_k~=~\frac{d}{h}~=~96$. To improve generalization and gradient flow in this deep architecture, we employ \textit{Stochastic Depth} (DropPath) \cite{Huang2016}. We linearly increase the drop probability $\rho_\ell$ across layers
\begin{equation}      
    \rho_\ell = \rho_{\max} \cdot \frac{\ell}{L}, \quad \text{with } \rho_{\max} = 0.1. 
\end{equation}
This randomly drops entire residual branches during training, forcing the network to learn redundant pathways for robust feature extraction. 

\textbf{Prior Decoder.}
The final token sequence $\mathbf{z}^{(L)}$ encodes the global vascular topology. To recover the spatial vessel map, we reshape the sequence back to the patch grid $\mathbf{Z}^{(L)} \in \mathbb{R}^{\frac{H}{p} \times \frac{W}{p} \times d}$.
Progressive upsampling via three transposed convolutional blocks restores full resolution, doubling spatial dimensions at each stage while reducing channels. 
A minimal prediction head produces global connectivity prior $P_A \in [0,1]^{H \times W}$ and guides subsequent U-Net refinement.

\begin{table}[ht]
\centering
\caption{Hyperparameters for the HAC architecture and training protocol. The Attention backbone learns global topology, while the U-Net performs local refinement.}
\label{tab:hac_params}
\setlength{\tabcolsep}{5pt}
\begin{tabular}{lc}
\toprule
\textbf{Parameter} & \textbf{Value} \\
\midrule
\multicolumn{2}{l}{\textit{\textbf{Attention Backbone ($A_{\psi}$)}}} \\
Input Resolution / Patch Size & $512 \times 512 \times  3$ / $8$ \\
Embed Dim ($d$) / MLP Ratio ($r$) & $384$ / $4.0$ \\
Depth ($L$) / Heads ($h$) & $8$ / $4$ \\
Dropout / DropPath & $0.1$ / $0.2$ \\
\midrule
\multicolumn{2}{l}{\textit{\textbf{U-Net Decoder ($U_{\Phi}$)}}} \\
Base Channels & $32$ \\
Encoder Scales & $[1, 2, 4, 8]$ \\
\midrule
\multicolumn{2}{l}{\textit{\textbf{Training Protocol}}} \\
Optimizer & AdamW ($10^{-4}$, Cosine) \\
Batch Size & $1$ \\
Warmup Epochs & $10$ \\
\multicolumn{2}{l}{\textit{\textbf{Stage 2}}} \\
Loss Weights ($\lambda$) & $T\text{:}1.5, cl\text{:}0.4, D\text{:}0.2, BCE\text{:}1.0$ \\
Tversky Params ($\alpha$ / $\beta$) & $0.2$ / $0.8$ \\
\multicolumn{2}{l}{\textit{\textbf{Stage 3}}} \\
Loss Weights ($\lambda$) & $T\text{:}2.0, cl\text{:}1.5$ \\
Tversky Params ($\alpha$ / $\beta$) & $0.1$ / $0.9$ \\
\bottomrule
\end{tabular}
\end{table}

\subsubsection{U-Net for Thin Vessel Edges}
Based on the vessel prior concatenated with the original input image, the U-Net now predicts the residual refinement map to recover thin peripheral branches.

\textbf{U-Net Architecture.}
The U-Net operates on a four-channel input formed by concatenating the RGB image with the attention network's vessel probability map
\begin{equation}
\mathbf{I}_{\text{aug}} = [I; P_{\text{A}}] \in \mathbb{R}^{H \times W \times 4},
\end{equation}
where global connectivity prior $P_{\text{A}}$ is the output from the attention network. The U-Net follows the standard encoder-decoder architecture with skip connections \cite{Ronneberger2015}. The encoder consists of four downsampling stages with feature dimensions $\{32, 64, 128, 256\}$, each comprising two convolutional layers with $3 \times 3$ kernels, batch normalization, ReLU activation, and $2 \times 2$ max pooling. 

The decoder mirrors this structure with transposed convolutions for upsampling and concatenation of encoder features via skip connections. The final layer produces a single-channel residual map $P_{\text{U}} \in [0,1]^{H \times W}$. This map represents the fine-scale correction needed to complete the segmentation. 

\bigskip
The final vessel probability map combining attention and convolution
\begin{equation}
P_{\text{HAC}} =  P_{\text{A}} + P_{\text{U}}
\end{equation}
is computed as the element-wise sum of the topological prior and the refinement residual. This additive design ensures that the global connectivity established by the attention backbone is preserved, while the U-Net focuses only on sharpening boundaries and adding missing thin vessels.

\subsection{Training Strategy}
To deal with the challenge of limited data availability in underexplored bladder vessel segmentation, we follow a three-stage training protocol designed to maximize data efficiency while ensuring domain robustness: 
\begin{enumerate}
    \item \textbf{Un-supervised physics-aware pre-training}: The model backbone is initialized by denoising physics-aware augmentations. 
    \item \textbf{Global topology learning}: The attention mechanism is optimized to capture the global vessel graph.
    \item \textbf{HAC refinement}: The CNN decoder is fine-tuned to resolve fine-scale vessel details using the learned topology.
\end{enumerate}

\smallskip
\subsubsection{Un-supervised physics-aware pre-training}
For cystoscopic videos, standard transfer learning often fails to capture the specific intraoperative- and geometry-dependent features. To address this gap, we introduce a self-supervised pre-training stage that forces the model to learn the underlying physical characteristics of the bladder environment. 

We formulate this as a denoising and restoration task, where the model must recover the original input image $I$ from a synthetically physics-aware corrupted version. 

\textbf{Physics-Aware Augmentations.} 
Rather than generic transformations, we apply three randomized augmentations $\tilde{I}$ that explicitly model cystoscopic challenges: (1) \textit{Synthetic Bubbles}, modeled as randomized elliptical occlusions, to mimic fluid turbulence and debris; (2) \textit{Photometric Variations}, including strong vignetting and variable gain, to simulate endoscopic light source instability; and (3) \textit{Local Contrast Reduction} to replicate motion blur and sensor noise.

\textbf{Training Objective.} The full HAC model is trained end-to-end to reconstruct the original image $I$ from $\tilde{I}$. This effectively initializes the feature extractors with robust, texture-aware representations before any semantic labels are introduced. Since the standard HAC output is a single-channel segmentation mask, we utilize a temporary reconstruction head to project the decoder features back to the RGB image space. The objective is to minimize the Mean Squared Error (MSE) 
\begin{equation} \mathcal{L}_{\mathrm{Stage1}} = || \tilde{P}_{\mathrm{HAC}}(\tilde{I}) - I ||^2 
\label{eq:loss_pretrain} 
\end{equation} 
between the reconstruction $\tilde{P}_{\mathrm{HAC}}$ and the clean signal $I$. 

\subsubsection{Global topology learning}
Since the attention backbone  $A_{\psi}$ should focus exclusively on global connections, the training labels must be adapted. These adaptions include the removal of all short vessel paths ($<100$~px) as well as the removal of all branches that lead to endpoints identified via 8-neighbor connectivity. With this optimized, topology-aware ground truth targets $M^*$, the attention backbone is trained in isolation from the U-Net, initialized with the weights from the physics-aware pre-training (Stage 1).

\textbf{Training Objective.} The attention network minimizes a composite loss against $M^*$
\begin{equation}
\label{eq:loss_attn}
\begin{aligned}
\mathcal{L}_{\mathrm{Stage2}}
&= \mathrm{BCE}(P_{\mathrm{A}}, M^*) 
+ \lambda_{\mathrm{D}} \mathcal{L}_{\mathrm{Dice}}(P_{\mathrm{A}}, M^*) \\
&\quad + \lambda_{\mathrm{cl}} \mathcal{L}_{\mathrm{clDice}}(P_{\mathrm{A}}, M^*) +\lambda_{\mathrm{T}} \mathcal{L}_{\mathrm{Tversky}}(P_{\mathrm{A}}, M^*)
\end{aligned}
\end{equation}
where $\mathcal{L}_{\mathrm{clDice}}$ \cite{Shit2021} penalizes centerline discontinuities, the Dice loss $\mathcal{L}_{\mathrm{Dice}}$  encourages spatial overlap, and the  Tversky loss $\mathcal{L}_{\mathrm{Tversky}}$ \cite{Abraham2019} suppresses non-vascular folds as false positives. Training is performed using AdamW with cosine annealing and a 10-epoch linear warmup.

\subsubsection{HAC refinement}
With the attention network $A_{\psi}$ frozen at its Stage~2 optimum, we train the U-Net decoder $U_{\Phi}$ to map from connectivity priors to precise segmentation masks. The U-Net now learns  to restore fine branches, sharp boundaries, and local details that were intentionally removed from the attention target $M^*$. Stage~3 minimizes the  composite loss 
\begin{equation}
\label{eq:loss_hac}
\begin{aligned}
\mathcal{L}_{\mathrm{Stage3}}
&= \lambda_{\mathrm{cl}} \mathcal{L}_{\mathrm{clDice}}(P_{\mathrm{HAC}}, G) +\lambda_{\mathrm{T}} \mathcal{L}_{\mathrm{Tversky}}(P_{\mathrm{HAC}}, G)
\end{aligned}
\end{equation}
with respect to the full ground truth annotations $G$, 
where $P_\mathrm{HAC} = \sigma(U_{\phi}([I; P_{\mathrm{A}}]))$ is the sigmoid-activated output from the joint HAC module. Table~\ref{tab:hac_params} shows all hyperparameters.

\begin{table*}[h]
    \centering
    \caption{Evaluation results (Accuracy, Dice, IoU, Precision, Recall, and clDice) comparing U-Net, U-Net++, Swin-UNet, SA-Net, and HAC~(Ours). ↑ indicates that higher values are better. The best and second best results are highlighted in boldface and underlined, respectively.}
    \label{tab:sota_comparison}
    \begin{tabular}{llcccccc}
        \toprule
        \textbf{Method} & \textbf{Type} & \textbf{Acc ↑} & \textbf{Dice ↑} & \textbf{IoU ↑} & \textbf{Prec ↑} & \textbf{Rec ↑} & \textbf{clDice ↑}\\
        \hline
        \multicolumn{8}{l}{\textit{CNN-based Methods}} \\
        \textbf{U-Net \cite{Ronneberger2015}} & Baseline CNN & \accunet & \diceunet & \iouunet & \precunet & \underline{\recunet} & \cldiceunet\\
        \textbf{UNet++ \cite{Zhou2018}} & + Dense skip & \accunetpp & \diceunetpp & \iouunetpp & \precunetpp & \textbf{\recunetpp} & \cldiceunetpp \\
        \hline
        \multicolumn{8}{l}{\textit{Transformer-based Methods}} \\
        \textbf{Swin-UNet \cite{Cao2022}} & Swin + U-Net & \textbf{\accswinunet} & \diceswinunet & \iouswinunet & \precswinunet & \recswinunet & \cldiceswinunet  \\
        \hline
        \multicolumn{8}{l}{\textit{Retinal Vessel Methods}} \\
        \textbf{SA-UNet \cite{Oktay2018}} & Retinal SOTA & \textbf{\accsaunet} & \textbf{\dicesaunet} & \textbf{\iousaunet} & \underline{\precsaunet} & \recsaunet & \underline{\cldicesaunet}  \\
        \hline
        \multicolumn{8}{l}{\textit{Ours}} \\
        \textbf{HAC} & ViT+CNN & \textbf{\acchac} & \underline{\dicehac} & \underline{\iouhac} & \textbf{\prechac} & \rechac & \textbf{\cldicehac}  \\
        \bottomrule
    \end{tabular}
\end{table*}

\subsection{Experiments}
All models are implemented in PyTorch 3.9 and trained using PyTorch Lightning 2.4. Batch size is set to 1. 

\textbf{Data Splits.}
For the pretraining (Stage~1), we extract 81 unlabeled frames from four separate cystoscopic videos. From the 50 annotated frames in BlaVeS \cite{Krauss2025}, we assign the same 90\% (45 frames) to the training stages 2 and 3, and use the  held-out set of 5 frames (10\%) for the test evaluation.

\section{Results}\label{sec:results}
After comparing HAC to state-of-the-art methods, we will emphasize its performance on endoscopic challenges using video frames from the BlaVeS dataset and an unseen video for temporal consistency.

\subsection{Comparison with State-of-the-Art}
We evaluate HAC against four established methods from \emph{retinal vessel segmentation} on the BlaVeS test set: 
\begin{enumerate}
	\item \textbf{U-Net} \cite{Ronneberger2015}, the standard baseline for biomedical segmentation
	\item \textbf{U-Net++} \cite{Zhou2018}, which introduces nested dense skip connections
	\item \textbf{Attention (SA) U-Net} \cite{Oktay2018}, which integrates attention gates to suppress background noise
    \item \textbf{Swin-U-Net} \cite{Cao2022}, that combines transformer encoders with CNN decoders
\end{enumerate}

To quantify their performance, we rely on standard pixel-wise segmentation metrics derived from True Positive (TP), True Negative (TN), False Positive (FP), and False Negative (FN) classifications including
\begin{equation} Accuracy=\frac{TP+TN}{TP+TN+FP+FN} \end{equation} 
that measures global correctness but can be misleading in class-imbalanced vessel segmentation, 
\begin{equation} Precision=\frac{TP}{TP+FP} \end{equation} 
that evaluates how many predicted vessel pixels are actually vessels, 
\begin{equation} Sensitivity=\frac{TP}{TP+FN} \end{equation} 
 that defines the proportion of actual vessel pixels correctly detected, 
\begin{equation} IoU=\frac{TP}{TP+FP+FN},  \end{equation} 
 and 
\begin{equation} Dice=\frac{2TP}{2TP+FP+FN} \end{equation} 
that assess the spatial overlap between the prediction and the ground truth. 
By calculating the overlap between the predicted segmentation and the skeleton of the ground truth, $clDice$ explicitly penalizes centerline discontinuities. Table~\ref{tab:sota_comparison} reports quantitative results.

HAC achieves an accuracy of \acchac, performing comparably to other leading models (Swin: \accswinunet~and SA-UNet: \accsaunet). In terms of spatial overlap, our model achieves competitive results, ranking second-highest in Dice coefficient (\dicehac) and IoU (\iouhac). Most importantly, HAC achieves the \emph{highest precision} (\prechac) among all compared methods. This metric is particularly relevant in cystoscopy, as it reflects the model's ability to minimize false positives and thereby effectively distinguish vascular structures from artifacts such as mucosal folds. Moreover, HAC excels in topological connectivity with a clDice score of \cldicehac~(vs. \cldicesaunet~for the next best method). This indicates the attention backbone’s role in preserving the global vessel network essential for navigation.

\begin{figure*}[ht!]
	\centering
    \def\svgwidth{\textwidth}
    \includegraphics[width=1\linewidth]{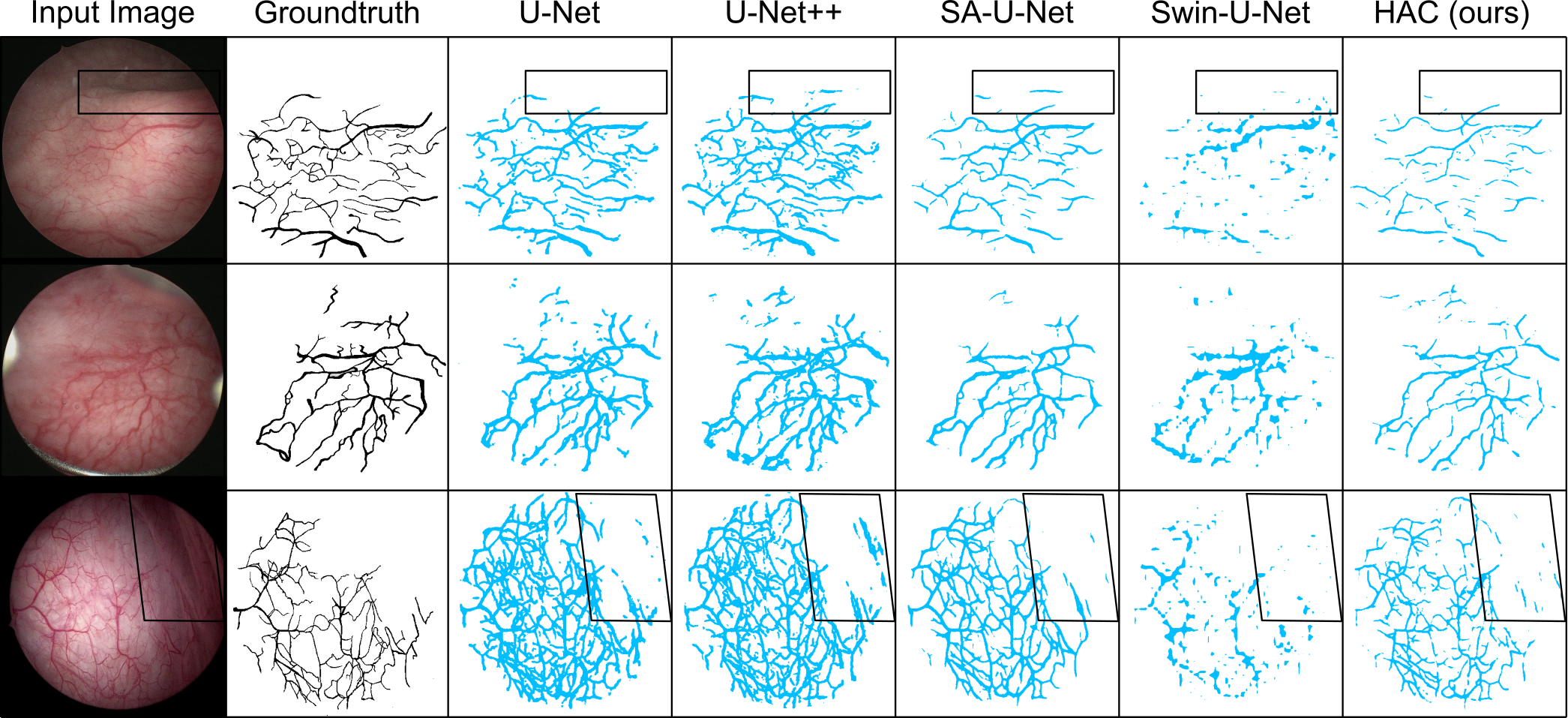}
	\caption{Qualitative comparison on challenging cystoscopic conditions using video frames from the BlaVes testset. Top:  Prominent tissue fold. Middle: Severe motion blur. Bottom: Strong vignetting and edge artifacts. Boxed regions highlight difficult areas.}
\label{fig:experiments_sota_blaves}
\end{figure*}

\subsection{Tackling Cystoscopic Challenges}
The primary difficulty with reliable intraoperative orientation in the bladder is the reliability of the segmented vascular map. Generalizability across interventions is particularly hindered by the over-segmentation of small, irrelevant vessels as well as mucosal folds that can be mistaken for blood vessels. 
Because these folds are not permanent anatomical features, their misclassification as vessels introduces ``false positive" landmarks, which makes them unsuitable for re-identification and can lead to errors in feature-based navigation.

\textbf{Robustness to Intraoperative Artifacts.} Apart from standard segmentation metrics, the reliability for intraoperative navigation is best demonstrated by the qualitative analysis of three test images in Figure~\ref{fig:experiments_sota_blaves}. These cystoscopic frames cover a wide range of colors, contrasts, and lighting conditions, demonstrating the difficulty of generalization. The different images highlight distinct difficulties. Image 1 (top) shows a large fold that must be excluded from segmentation.  Image 2 (middle) is generally blurry and has low contrast. Image 3 (bottom) shows folds on the right side with characteristics similar to those of vessels. 

While nearly every model struggles to avoid predicting the fold in the first image, only our model and the U-Net do not predict anything there. This is reflected by our high precision without losing correct predictions (maintaining high accuracy and Dice). 

In contrast, U-Net predicts numerous small vessels that are irrelevant to navigation. In the second image, the challenge is to segment the thin vessels instead of blurry, unclear connections; here, SA-UNet and HAC perform the best by predicting only the prominent vessels as thin, clear structures. Finally, in the last image, the focus again lies on non-identifying folds while still accurately predicting vessels. Even though our model still predicts some tiny structures in the fold area, they are significantly smaller than in the other models (except Swin-U-Net, which does not perform well at all), where the folds are incorrectly segmented as prominent vessel branches.

\begin{figure*}[ht!]
	\centering
    \def\svgwidth{\textwidth}
    \includegraphics[width=1\linewidth]{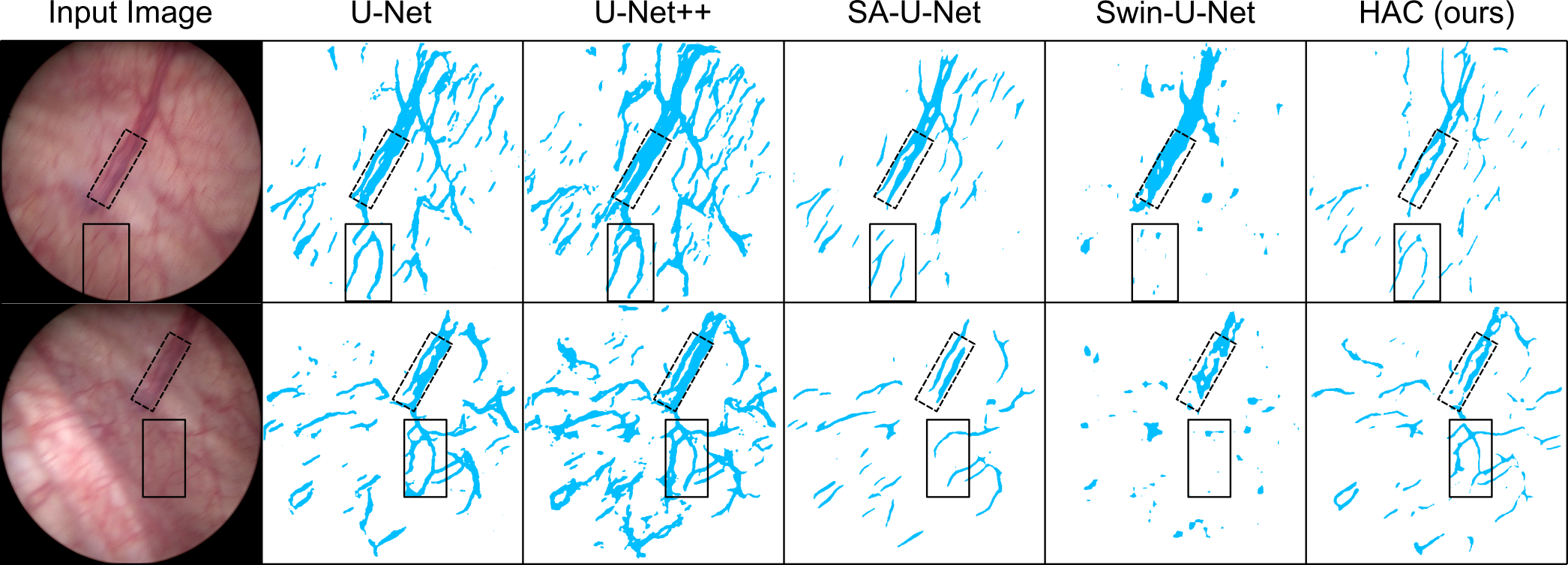}
	\caption{Qualitative comparison of vessel segmentation on two consecutive frames from the same cystoscopy. The boxed regions highlight the same anatomical area that must be matched across frames despite challenges such as motion blur, deformation, and a shifting field of view.} 
\label{fig:experiments_sota_different_angles}
\end{figure*}

\textbf{Temporal Consistency in Video Sequences.}
While the previous section addressed static, frame-level analysis, comparing frames within a continuous video sequence is most relevant for the clinical application.
In this context, maximizing total vessel recall is secondary to ensuring structural stability. For navigation, it is more important to segment reliable landmarks than to produce a noisy map. Figure~\ref{fig:experiments_sota_different_angles} presents frames from a surgery entirely excluded from both the supervised and self-supervised training stages. More precisely, the selected frames are two consecutive frames from the same surgery, and illustrate how the movement of the endoscope and the bladder dynamics add the challenges of motion blur, deformation, and a shifting field of view. Areas highlighted in the boxes correspond to the same area in the bladder and need to be matched across frames.
Contrary to its high quantitative metrics, SA-U-Net is not able to capture this consistency at all; the two structures are either interrupted or unrecognizable. In contrast, U-Net and U-Net++ suffer from over-segmentation: distinct structures (upper box) merge into one single large vessel due to noise. Conversely, our model predicts two distinct, connected vessels in the upper region and preserves the structure in the lower region, confirming the temporal stability essential for reliable intraoperative navigation.

\section{Discussion}\label{sec:discussion}

While the segmentation metrics, particularly the increased precision, indicate that our model performs well for this task, these differences must be interpreted cautiously given the \textbf{small test set} size and the inherent challenges of manual annotation in cystoscopic imaging. The limited number of hand-labeled frames (50 total) makes definitive performance ranking difficult based on pixel-level accuracy alone. 
However, the test set was carefully put together to maximize the qualitative diversity in images, featuring different failure modes such as strong mucosal folds, motion blur, and vignetting (see Figure~\ref{fig:experiments_sota_blaves}). Nevertheless, the metrics validate our hypothesis that methods optimized for controlled retinal imaging do not transfer to cystoscopy's unique challenges such as multi-layer vessels, variable illumination, or low color contrast. 
Notably, none of the models achieve the Dice and recall results that are common in the retinal field.

Additionally, it appears that each baseline model specializes in \textbf{different sub-tasks}: U-Net and U-Net++ capture even small vessels but fail to differentiate between navigationally important structures and noise. Consequently, while some structures are recognizable, the map is often misleading due to wrongly segmented folds or an overload of unimportant vessel fragments. SA-UNet, in contrast, excels in Dice score but fails to exclude folds from the segmentation and proves incapable of consistently segmenting the same structures from different angles.

Our model effectively combines these advantages by maintaining sufficient sensitivity (via the U-Net component) while reasoning about general structure (via the Attention component). As a result, while it does not achieve the highest numbers in all metrics, \emph{it excels in the most critical one} for intraoperative navigation within the bladder: it almost entirely avoids segmenting folds and consistently predicts recognizable, reliable structures.

\section{Conclusion}\label{sec:conclusion}
While blood vessels visible during endoscopic surgery are the only reliable and deformation-invariant landmarks in the urinary bladder, they are also intuitive and clinically interpretable. Hence, a robust segmentation of the vascular structures is essential to create a 
digital twin of the bladder and offer intraoperative guidance. Yet, this field of research is so far limited by the amount, but also quality of available data, especially due to variations in illumination, severe organ deformation, or the superposition of vascular layers.
Not surprisingly, existing vessel segmentation methods often fail under these conditions, resulting in either missed structures or the over-segmentation of artifacts.
In this work, we tackled these challenges by introducing the Hybrid Attention-Convolution (HAC) framework, which combines an attention backbone to capture global topology with a U-Net that predicts residual details. We further robustified this approach through physics-aware pretraining, which uses clinically grounded augmentations to master cystoscopic physical characteristics. Our method demonstrates superior precision and temporal stability, successfully distinguishing true vasculature from geometric mimics to enable reliable intraoperative navigation.
In a next step, we will use these segmented vessels to extract stable features for matching across frames. This step is essential for building a complete vascular map of the bladder, towards a \emph{digital twin with patient-specific vascular fingerprint}.

\bigskip
\appendix

\setlength{\tabcolsep}{6pt} 
\renewcommand{\arraystretch}{1.1} 
\begin{table*}[t!]
    \centering
    \caption{\textbf{Quantitative comparison of BlaVeS and DRIVE} across key vessel segmentation metrics. Arrows indicate whether higher values simplify (↑) or complicate (↓) the segmentation task.  "Ratio" = the percentage of pixels in each class (vessel, background, or overall FOV). RI = red intensity, RCC = red channel contrast, CSI = color separation index, TI = tortuosity index, BD = branching density, CV = coefficient of variation, VI = vignetting index. All intensity metrics computed on 8-bit images (0–255). RCC, TI, CV, and VI are unitless. BD is reported per 100 px of skeleton length. Bold numbers indicate easier segmentation.}
   \label{tab:dataset}
    \begin{tabular}{l@{\hspace{5pt}}c@{\hspace{5pt}}c@{\hspace{5pt}}c@{\hspace{5pt}}c@{\hspace{5pt}}c@{\hspace{5pt}}c@{\hspace{5pt}}c@{\hspace{5pt}}cc}
        \toprule
        \multicolumn{1}{c}{} & \multicolumn{1}{c}{} 
            & \multicolumn{3}{c}{\textbf{Superposition of Vascular Layers}}
            & \multicolumn{2}{c}{\textbf{Tortuous Vasculature}} 
            & \multicolumn{2}{c}{\textbf{Illumination}} \\
        \cmidrule(lr){3-5} \cmidrule(lr){6-7} \cmidrule(lr){8-9}
        \textbf{Dataset} & \textbf{Ratio} in \% 
            & \textbf{RI} [0,255] & \textbf{RCC} & \textbf{CSI ↑}
            & \textbf{TI ↓} & \textbf{BD ↓} 
            & \textbf{CV ↓} & \textbf{VI ↓} \\
        \midrule
         \textbf{BlaVeS}    & \meanstd{\blavesOverallRatioMean}{\blavesOverallRatioStd}
                            & \meanstd{\blavesOverallRIMean}{\blavesOverallRIStd}
                            & \meanstd{\blavesOverallRCCMean}{\blavesOverallRCCStd}
                            & \meanstd{\textbf{\blavesOverallCSIMean}}{\blavesOverallCSIStd}
                            & 
                            & 
                            & \meanstd{\blavesOverallCVMean}{\blavesOverallCVStd}
                            & \meanstd{\blavesOverallVIMean}{\blavesOverallVIStd}
                            \\
        \textbf{DRIVE}     & \meanstd{\driveOverallRatioMean}{\driveOverallRatioStd}
                            & \meanstd{\driveOverallRIMean}{\driveOverallRIStd}
                            & \meanstd{\driveOverallRCCMean}{\driveOverallRCCStd}
                            & \meanstd{\driveOverallCSIMean}{\textbf{\driveOverallCSIStd}}
                            & 
                            & 
                            & \textbf{\meanstd{\driveOverallCVMean}{\driveOverallCVStd}}
                            & \textbf{\meanstd{\driveOverallVIMean}{\driveOverallVIStd}}
                            \\
        \midrule
        \textbf{Vessel BlaVeS} & \meanstd{\blavesVesselRatioMean}{\blavesVesselRatioStd}
                          & \meanstd{\blavesVesselRIMean}{\blavesVesselRIStd}
                          &
                          & 
                          & \meanstd{\blavesVesselTIMean}{\blavesVesselTIStd}
                          & \meanstd{\blavesVesselBDMean}{\blavesVesselBDStd}
                          & \meanstd{\blavesVesselCVMean}{\blavesVesselCVStd}
                          & \meanstd{\blavesVesselVIMean}{\blavesVesselVIStd} \\
        \textbf{Vessel DRIVE}  & \meanstd{\driveVesselRatioMean}{\driveVesselRatioStd}
                          & \meanstd{\driveVesselRIMean}{\driveVesselRIStd}
                          & 
                          & 
                          & \textbf{\meanstd{\driveVesselTIMean}{\driveVesselTIStd}}
                          & \textbf{\meanstd{\driveVesselBDMean}{\textbf{\driveVesselBDStd}}}
                          & \textbf{\meanstd{\driveVesselCVMean}{\driveVesselCVStd}}
                          & \textbf{\meanstd{\driveVesselVIMean}{\driveVesselVIStd}} \\
        \midrule
        \textbf{BG BlaVeS} & \meanstd{\blavesBGRatioMean}{\blavesBGRatioStd}
                           & \meanstd{\blavesBGRIMean}{\blavesBGRIStd}
                           & 
                           &
                           & 
                           & 
                           & \meanstd{\blavesBGCVMean}{\blavesBGCVStd}
                           & \meanstd{\blavesBGVIMean}{\blavesBGVIStd} \\
        \textbf{BG DRIVE}  & \meanstd{\driveBGRatioMean}{\driveBGRatioStd}
                           & \meanstd{\driveBGRIMean}{\driveBGRIStd}
                           & 
                           & 
                           & 
                           & 
                           &\textbf{ \meanstd{\driveBGCVMean}{\driveBGCVStd}}
                           & \textbf{\meanstd{\driveBGVIMean}{\driveBGVIStd}} \\   
         \bottomrule        
    \end{tabular}
\end{table*}
\setlength{\tabcolsep}{6pt}

Systematic image‑quality profiling of BlaVeS highlights multiple segmentation-relevant challenges, summarized in Table~\ref{tab:dataset}. The dataset is highly unbalanced, with only \meanstdpercent{\blavesVesselRatioMean}{\blavesVesselRatioStd} vessel pixels. For all subsequent analysis, "vessel" (indicated as "v") refers to annotated vessel regions, while "background" (indicated as "bg") refers only to bladder-wall tissue and excludes the black borders surrounding the endoscopic field of view (FOV).
\subsubsection{Vascular Layer Superposition}
Although vessels are often expected to exhibit higher red-channel intensity than the surrounding mucosa, BlaVeS shows only a small difference in mean red intensity values (\meanstd{\blavesVesselRIMean}{\blavesVesselRIStd} vs. \meanstd{\blavesBGRIMean}{\blavesBGRIStd}), indicating limited discriminability by redness alone.
To quantify this, the Red Channel Contrast is defined as 
\begin{equation}
	\mathrm{RCC} = \frac{R_{\mathrm{v}} - R_{\mathrm{bg}}}{R_{\mathrm{bg}}}, 
\end{equation} 
where $R_{\mathrm{v}}$ and $R_{\mathrm{bg}}$ denote the mean red intensities of vessel and background pixels, respectively. 
Although retinal vessels typically appear darker than surrounding tissue due to blood absorption (DRIVE: $\mathrm{RCC}=$\meanstd{\driveOverallRCCMean}{\driveOverallRCCStd}), bladder vessels do not exhibit the strong red appearance one might expect from hemoglobin-rich tissue. The minimal red-channel contrast (BlaVeS: $\mathrm{RCC}=$\meanstd{\blavesOverallRCCMean}{\blavesOverallRCCStd}) indicates that vessels and mucosa have nearly identical red intensities, providing limited color-based discriminability. The Color Separation Index
\begin{equation}
	\mathrm{CSI}
	= \left\lVert \boldsymbol{\mu}_\mathrm{v} - \boldsymbol{\mu}_{\mathrm{bg}} \right\rVert_2,
	\qquad \boldsymbol{\mu}\in\mathbb{R}^3, 
\end{equation} 
with ${\mu}$ the mean RGB vectors, measures the Euclidean distance in RGB color space between the mean vessel and background intensities.  
While BlaVeS shows comparable mean CSI to DRIVE (\meanstd{\blavesOverallCSIMean}{\blavesOverallCSIStd} vs. \meanstd{\driveOverallCSIMean}{\driveOverallCSIStd}), the nearly 3-fold larger standard deviation reveals high inter-frame variability: some frames exhibit strong color separation, while others show near-identical vessel and background colors. This inconsistency undermines color-based segmentation cues, requiring models to rely more heavily on texture and geometric features.

\subsubsection{Tortuous Vasculature}
The tortuosity of vessels in BlaVeS was evaluated using two complementary metrics. The Tortuosity Index 
\begin{equation}
	\mathrm{TI} = \frac{L_{\mathrm{path}}}{d_{\mathrm{euc}}}, 
\end{equation}
is defined as the ratio of vessel path length $L_{\mathrm{path}}$ to the Euclidean distance between endpoints $d_{\mathrm{euc}}$. BlaVeS shows a mean TI of \meanstd{\blavesVesselTIMean}{\blavesVesselTIStd}, indicating that most vessels are not strongly curved when considered over their entire length. Branching complexity is quantified through Branching Density 
\begin{equation}
	\mathrm{BD} = \frac{\mathrm{\#~branch~points}}{\mathrm{L_\mathrm{path}}}.
\end{equation}
The mean $\mathrm{BD=}$\meanstd{\blavesVesselBDMean}{\blavesVesselBDStd} (per 100 px) is close to retinal values (\meanstd{\driveVesselBDMean}{\driveVesselBDStd}) but shows much higher variability. This reflects non-uniform branching, where some frames contain dense junction-rich regions while others are relatively sparse. Such variability increases segmentation difficulty, especially in areas of multi-layer overlap, where branches can differ greatly in color and contrast. In these regions, junctions may be partially blurred, low in separability from the background, or distorted by specular highlights, leading to ambiguous or spurious connections. 

\subsubsection{Illumination}
Endoscopic imaging introduces pronounced nonuniform lighting due to point-source illumination and varying tissue-camera distances. To quantify spatial brightness variation, each image is divided into a 3×3 grid (covering only the field of view), and the Coefficient of Variation
\begin{equation}
	\mathrm{CV} = \frac{\sigma_{\mathrm{blocks}}}{\mu_{\mathrm{blocks}}}
\end{equation}
is computed as the ratio of the standard deviation $\sigma_{\mathrm{blocks}}$ to the mean of block-wise intensities $\mu_{\mathrm{blocks}}$. BlaVeS shows $\mathrm{CV} =$ \meanstd{\blavesOverallCVMean}{\blavesOverallCVStd}, substantially higher than DRIVE’s \meanstd{\driveOverallCVMean}{\driveOverallCVStd}, indicating stronger illumination gradients across the field of view. The vignetting index 
\begin{equation}
	\mathrm{VI} = \frac{I_{\mathrm{c}}-I_{\mathrm{p}}}{I_{\mathrm{c}}}
\end{equation}
measures center-to-periphery intensity drop, where $I_{\mathrm{c}}$ and $I_{\mathrm{p}}$ are the mean intensities in the central and peripheral image regions, respectively. BlaVeS exhibits $\mathrm{VI}=$ \meanstd{\blavesOverallVIMean}{\blavesOverallVIStd}, with high variability across frames, reflecting inconsistent vignetting patterns between recordings.

These quantitative findings demonstrate that cystoscopic vessel segmentation presents fundamentally different challenges from retinal imaging: weak and inconsistent color contrast, multi-layer depth ambiguity, and highly variable illumination. These factors necessitate architectural designs that integrate global context with local refinement, as addressed in Section~\ref{sec:methods}.

\section*{References}
\bibliography{main} 

\begin{thebibliography}{10}
\providecommand{\url}[1]{#1}
\csname url@samestyle\endcsname
\providecommand{\newblock}{\relax}
\providecommand{\bibinfo}[2]{#2}
\providecommand{\BIBentrySTDinterwordspacing}{\spaceskip=0pt\relax}
\providecommand{\BIBentryALTinterwordstretchfactor}{4}
\providecommand{\BIBentryALTinterwordspacing}{\spaceskip=\fontdimen2\font plus
\BIBentryALTinterwordstretchfactor\fontdimen3\font minus
  \fontdimen4\font\relax}
\providecommand{\BIBforeignlanguage}[2]{{%
\expandafter\ifx\csname l@#1\endcsname\relax
\typeout{** WARNING: IEEEtran.bst: No hyphenation pattern has been}%
\typeout{** loaded for the language `#1'. Using the pattern for}%
\typeout{** the default language instead.}%
\else
\language=\csname l@#1\endcsname
\fi
#2}}
\providecommand{\BIBdecl}{\relax}
\BIBdecl

\bibitem{Jubber2023}
I.~Jubber, S.~Ong, L.~Bukavina, P.~C. Black, E.~Compérat, A.~M. Kamat,
  L.~Kiemeney, N.~Lawrentschuk, S.~P. Lerner, J.~J. Meeks, H.~Moch, A.~Necchi,
  V.~Panebianco, S.~S. Sridhar, A.~Znaor, J.~W. Catto, and M.~G. Cumberbatch,
  ``Epidemiology of bladder cancer in 2023: a systematic review of risk
  factors,'' \emph{European urology}, vol.~84, no.~2, pp. 176--190, 2023.

\bibitem{Abbas2025}
S.~Abbas, R.~Shafik, N.~Soomro, R.~Heer, and K.~Adhikari, ``Ai predicting
  recurrence in non-muscle-invasive bladder cancer: systematic review with
  study strengths and weaknesses,'' \emph{Frontiers in oncology}, vol.~14, p.
  1509362, 2025.

\bibitem{Lubbad2024}
M.~Lubbad, D.~Karaboga, A.~Basturk, B.~Akay, U.~Nalbantoglu, and
  I.~Pa{\c{c}}al, ``Machine learning applications in detection and diagnosis of
  urology cancers: a systematic literature review,'' \emph{Neural Computing and
  Applications}, vol.~36, no.~12, pp. 6355--6379, 2024.

\bibitem{Sudhi2023}
\BIBentryALTinterwordspacing
M.~Sudhi, V.~K. Shukla, D.~K. Shetty, V.~Gupta, A.~S. Desai, N.~Naik, and B.~Z.
  Hameed, ``Advancements in bladder cancer management: A comprehensive review
  of artificial intelligence and machine learning applications,''
  \emph{Engineered Science}, vol.~26, p. 1003, 2023. [Online]. Available:
  \url{http://dx.doi.org/10.30919/es1003}
\BIBentrySTDinterwordspacing

\bibitem{Schuele2022a}
J.~Sch{\"u}le, A.~Salehah, P.~Somers, N.~Harland, C.~Tar{\'\i}n, A.~Stenzl, and
  O.~Sawodny, ``Real-time vascular graph extraction for surgical navigation,''
  2022.

\bibitem{Krauss2025}
F.~Krauß, N.~Smati, M.~Ege, Z.~Lovasz, O.~Sawodny, and C.~Veil, ``Blaves: A
  novel hand-labeled dataset for improved bladder vessel segmentation with
  modified u-net,'' \emph{47th Annual International Conference of the IEEE
  Engineering in Medicine and Biology Society}, 2025.

\bibitem{Staal2004}
J.~Staal, M.~Abramoff, M.~Niemeijer, M.~Viergever, and B.~van Ginneken,
  ``Ridge-based vessel segmentation in color images of the retina,'' \emph{IEEE
  Transactions on Medical Imaging}, vol.~23, no.~4, pp. 501--509, April 2004.

\bibitem{Hoover2000}
A.~Hoover, V.~Kouznetsova, and M.~Goldbaum, ``Locating blood vessels in retinal
  images by piecewise threshold probing of a matched filter response,''
  \emph{IEEE Transactions on Medical Imaging}, vol.~19, no.~3, pp. 203--210,
  2000.

\bibitem{Fraz2012}
M.~M. Fraz, P.~Remagnino, A.~Hoppe, B.~Uyyanonvara, A.~R. Rudnicka, C.~G. Owen,
  and S.~A. Barman, ``An ensemble classification-based approach applied to
  retinal blood vessel segmentation,'' \emph{IEEE Transactions on Biomedical
  Engineering}, vol.~59, no.~9, pp. 2538--2548, Sep. 2012.

\bibitem{Chaudhuri1989}
S.~Chaudhuri, S.~Chatterjee, N.~Katz, M.~Nelson, and M.~Goldbaum, ``Detection
  of blood vessels in retinal images using two-dimensional matched filters,''
  \emph{IEEE Transactions on medical imaging}, vol.~8, no.~3, pp. 263--269,
  1989.

\bibitem{Zana2001}
F.~Zana and J.-C. Klein, ``Segmentation of vessel-like patterns using
  mathematical morphology and curvature evaluation,'' \emph{IEEE transactions
  on image processing}, vol.~10, no.~7, pp. 1010--1019, 2001.

\bibitem{Soares2006}
J.~V. Soares, J.~J. Leandro, R.~M. Cesar, H.~F. Jelinek, and M.~J. Cree,
  ``Retinal vessel segmentation using the 2-d gabor wavelet and supervised
  classification,'' \emph{IEEE Transactions on medical Imaging}, vol.~25,
  no.~9, pp. 1214--1222, 2006.

\bibitem{Ronneberger2015}
O.~Ronneberger, P.Fischer, and T.~Brox, ``U-net: Convolutional networks for
  biomedical image segmentation,'' in \emph{Medical Image Computing and
  Computer-Assisted Intervention (MICCAI)}, ser. LNCS, vol. 9351.\hskip 1em
  plus 0.5em minus 0.4em\relax Springer, 2015, pp. 234--241.

\bibitem{Alom2018}
M.~Z. Alom, C.~Yakopcic, M.~Hasan, T.~M. Taha, and V.~K. Asari, ``Recurrent
  residual u-net for medical image segmentation,'' \emph{Journal of medical
  imaging}, vol.~6, no.~1, pp. 014\,006--014\,006, 2019.

\bibitem{Zhou2018}
Z.~Zhou, M.~M. Rahman~Siddiquee, N.~Tajbakhsh, and J.~Liang, ``{UNet++: A
  nested U-Net architecture for medical image segmentation},'' in \emph{DLMIA},
  2018.

\bibitem{Oktay2018}
O.~Oktay, J.~Schlemper, L.~L. Folgoc, M.~Lee, M.~Heinrich, K.~Misawa, K.~Mori,
  S.~McDonagh, N.~Y. Hammerla, B.~Kainz, B.~Glocker, and D.~Rueckert,
  ``Attention u-net: Learning where to look for the pancreas,'' \emph{arXiv
  preprint arXiv:1804.03999}, 2018.

\bibitem{Shit2021}
S.~Shit, J.~C. Paetzold, A.~Sekuboyina, I.~Ezhov, A.~Unger, A.~Zhylka, J.~P.
  Pluim, U.~Bauer, and B.~H. Menze, ``{clDice—a novel topology-preserving
  loss function for tubular structure segmentation},'' in \emph{CVPR}, 2021.

\bibitem{Nasr-Esfahani2016}
E.~Nasr-Esfahani, N.~Karimi, M.~H. Jafari, S.~M.~R. Soroushmehr, S.~Samavi,
  B.~K. Nallamothu, and K.~Najarian, ``{Segmentation of vessels in angiograms
  using convolutional neural networks},'' \emph{Biomedical Signal Processing
  and Control}, vol.~40, pp. 240--251, 2016.

\bibitem{Ajam2017}
A.~Ajam, A.~A. Aziz, V.~S. Asirvadam, A.~S. Muda, I.~Faye, and S.~J.
  Safdar~Gardezi, ``A review on segmentation and modeling of cerebral
  vasculature for surgical planning,'' \emph{IEEE Access}, vol.~5, pp.
  15\,222--15\,240, 2017.

\bibitem{Woo2018}
S.~Woo, J.~Park, J.-Y. Lee, and I.~S. Kweon, ``{CBAM: Convolutional block
  attention module},'' in \emph{ECCV}, 2018.

\bibitem{Dosovitskiy2020}
A.~Dosovitskiy, ``An image is worth 16x16 words: Transformers for image
  recognition at scale,'' \emph{arXiv preprint arXiv:2010.11929}, 2020.

\bibitem{Chen2021}
J.~Chen, Y.~Lu, Q.~Yu, X.~Luo, E.~Adeli, Y.~Wang, L.~Lu, A.~L. Yuille, and
  Y.~Zhou, ``Transunet: Transformers make strong encoders for medical image
  segmentation,'' \emph{arXiv preprint arXiv:2102.04306}, 2021.

\bibitem{Cao2022}
H.~Cao, Y.~Wang, J.~Chen, D.~Jiang, X.~Zhang, Q.~Tian, and M.~Wang,
  ``Swin-unet: Unet-like pure transformer for medical image segmentation,'' in
  \emph{European conference on computer vision}.\hskip 1em plus 0.5em minus
  0.4em\relax Springer, 2022, pp. 205--218.

\bibitem{Wang2022}
R.~Wang, T.~Lei, R.~Cui, B.~Zhang, H.~Meng, and A.~K. Nandi, ``Medical image
  segmentation using deep learning: A survey,'' \emph{IET image processing},
  vol.~16, no.~5, pp. 1243--1267, 2022.

\bibitem{Hatamizadeh2022}
A.~Hatamizadeh, Y.~Tang, V.~Nath, D.~Yang, A.~Myronenko, B.~Landman, H.~R.
  Roth, and D.~Xu, ``Unetr: Transformers for 3d medical image segmentation,''
  in \emph{Proceedings of the IEEE/CVF winter conference on applications of
  computer vision}, 2022, pp. 574--584.

\bibitem{Vaswani2017}
A.~Vaswani, ``Attention is all you need,'' \emph{Advances in Neural Information
  Processing Systems}, 2017.

\bibitem{Petit2021}
O.~Petit, N.~Thome, C.~Rambour, L.~Themyr, T.~Collins, and L.~Soler, ``U-net
  transformer: Self and cross attention for medical image segmentation,'' in
  \emph{Machine Learning in Medical Imaging: 12th International Workshop, MLMI
  2021, Held in Conjunction with MICCAI 2021, Strasbourg, France, September 27,
  2021, Proceedings 12}.\hskip 1em plus 0.5em minus 0.4em\relax Springer, 2021,
  pp. 267--276.

\bibitem{Chen2020}
T.~Chen, S.~Kornblith, M.~Norouzi, and G.~Hinton, ``A simple framework for
  contrastive learning of visual representations,'' in \emph{International
  conference on machine learning}.\hskip 1em plus 0.5em minus 0.4em\relax PmLR,
  2020, pp. 1597--1607.

\bibitem{He2022}
K.~He, X.~Chen, S.~Xie, Y.~Li, P.~Doll{\'a}r, and R.~Girshick, ``Masked
  autoencoders are scalable vision learners,'' in \emph{Proceedings of the
  IEEE/CVF conference on computer vision and pattern recognition}, 2022, pp.
  16\,000--16\,009.

\bibitem{Laine2017}
S.~Laine and T.~Aila, ``Temporal ensembling for semi-supervised learning,''
  2016.

\bibitem{Zhou2021}
H.-Y. Zhou, S.~Yu, C.~Bian, Y.~Hu, K.~Ma, and Y.~Zheng, ``Comparing to learn:
  Surpassing imagenet pretraining on radiographs by comparing image
  representations,'' in \emph{International Conference on Medical Image
  Computing and Computer-Assisted Intervention}.\hskip 1em plus 0.5em minus
  0.4em\relax Springer, 2020, pp. 398--407.

\bibitem{Azizi2021}
S.~Azizi, B.~Mustafa, F.~Ryan, Z.~Beaver, J.~Freyberg, J.~Deaton, A.~Loh,
  A.~Karthikesalingam, S.~Kornblith, T.~Chen \emph{et~al.}, ``Big
  self-supervised models advance medical image classification,'' pp.
  3478--3488, 2021.

\bibitem{Tobin2017}
J.~Tobin, R.~Fong, A.~Ray, J.~Schneider, W.~Zaremba, and P.~Abbeel, ``Domain
  randomization for transferring deep neural networks from simulation to the
  real world,'' in \emph{2017 IEEE/RSJ international conference on intelligent
  robots and systems (IROS)}.\hskip 1em plus 0.5em minus 0.4em\relax IEEE,
  2017, pp. 23--30.

\bibitem{Buslaev2020}
A.~Buslaev, V.~I. Iglovikov, E.~Khvedchenya, A.~Parinov, M.~Druzhinin, and
  A.~A. Kalinin, ``Albumentations: fast and flexible image augmentations,''
  \emph{Information}, vol.~11, no.~2, p. 125, 2020.

\bibitem{Blaves25}
\BIBentryALTinterwordspacing
F.~Krauß, ``{BlaVeS: Bladder Vessel Segmentation},'' 2025. [Online].
  Available: \url{https://doi.org/10.18419/DARUS-4763}
\BIBentrySTDinterwordspacing

\bibitem{Miodonski1999}
A.~J. Miodo{\'n}ski and J.~A. Litwin, ``Microvascular architecture of the human
  urinary bladder wall: a corrosion casting study,'' \emph{The Anatomical
  Record: An Official Publication of the American Association of Anatomists},
  vol. 254, no.~3, pp. 375--381, 1999.

\bibitem{Andersson2017}
K.-E. Andersson, D.~B. Boedtkjer, and A.~Forman, ``The link between vascular
  dysfunction, bladder ischemia, and aging bladder dysfunction,''
  \emph{Therapeutic advances in urology}, vol.~9, no.~1, pp. 11--27, 2017.

\bibitem{Chang2024}
S.~Chang, G.~A. Wintergerst, C.~Carlson, H.~Yin, K.~R. Scarpato, A.~N.
  Luckenbaugh, S.~S. Chang, S.~Kolouri, and A.~K. Bowden, ``Low-cost and
  label-free blue light cystoscopy through digital staining of white light
  cystoscopy videos,'' \emph{Communications Medicine}, vol.~4, no.~1, p. 269,
  2024.

\bibitem{Huang2016}
G.~Huang, Y.~Sun, Z.~Liu, D.~Sedra, and K.~Q. Weinberger, ``Deep networks with
  stochastic depth,'' in \emph{European conference on computer vision}.\hskip
  1em plus 0.5em minus 0.4em\relax Springer, 2016, pp. 646--661.

\bibitem{Abraham2019}
N.~Abraham and N.~M. Khan, ``A novel focal tversky loss function with improved
  attention u-net for lesion segmentation,'' in \emph{2019 IEEE 16th
  International Symposium on Biomedical Imaging (ISBI 2019)}, April 2019, pp.
  683--687.

\end{thebibliography}
\bibliographystyle{IEEEtran}

\end{document}